\documentclass{article}
\usepackage{spconf,amsmath,graphicx}
		
	\usepackage{float}
	\restylefloat{table}
	\usepackage{pseudocode}
	\usepackage{algorithmicx}
	\usepackage{epsfig}
	\usepackage{pdfpages}
	\usepackage{graphicx}
	\usepackage{multirow}
	\usepackage{amsmath,graphicx}
	\usepackage{amssymb}
	\usepackage{subfigure}
	\usepackage{color}
	 \usepackage{sidecap}
	\usepackage{amsthm}
	\newtheorem{thm}{Theorem}

	\theoremstyle{definition}
	
	\theoremstyle{remark}

\def\bea{\begin{eqnarray}}
\def\beas{\begin{eqnarray*}}
\def\eea{\end{eqnarray}}
\def\eeas{\end{eqnarray*}}

	\def\be{\begin{equation}}
	\def\bmat{\begin{matrix}}
	\def\emat{\end{matrix}}
	\def\bea{\begin{eqnarray}}
	\def\beas{\begin{eqnarray*}}
	\def\eea{\end{eqnarray}}
	\def\eeas{\end{eqnarray*}}

	\def\bi{\begin{itemize}}
	\def\ee{\end{equation}}
	\def\ei{\end{itemize}}

	\def\z1{z^{-1}}

	\def\bmat{\begin{matrix}}
	\def\emat{\end{matrix}}

\title{Two step recovery of jointly sparse and low-rank matrices: theoretical guarantees}

\name{Sampurna Biswas, Sunrita Poddar, Soura Dasgupta, Raghuraman Mudumbai, and Mathews Jacob \thanks{This work is in part  supported by US NSF grants  EPS-1101284, ECCS-1150801, CNS-1329657, CCF-1302456,  CCF-1116067, NIH 1R21HL109710-01A1, ACS RSG-11-267-01-CCE,  and ONR grant N00014-13-1-0202.}}
\address{Department of Electrical and Computer Engineering\\ The University of Iowa, IA, USA}

\begin{document}

\maketitle
	\textit{Abstract:} \textbf{We introduce a two step algorithm with theoretical guarantees to recover a jointly sparse and low-rank matrix from undersampled measurements of its columns. The algorithm first estimates the row subspace of the matrix using a set of common measurements of the columns. In the second step, the subspace aware recovery of the matrix is solved using a simple least square algorithm. The results are verified in the context of recovering CINE data from undersampled measurements; we obtain good recovery when the sampling conditions are satisfied.  } \\
\section{Introduction} 
The recovery of matrices that are simultaneously low-rank and jointly sparse from few measurements has received considerable attention in the recent years, mainly in the context of the of dynamic MRI reconstruction \cite{liang,lingala2011accelerated}. In this context, the columns of the matrix correspond to vectorized image frames, while the rows are the temporal profiles of each voxel. While there is considerable theoretical progress in problems such as recovering jointly sparse vectors or low-rank matrices, the recovery of matrices that are simultaneously low-rank and jointly sparse have received considerably less attention. 

Recently in \cite{golbabaee2012compressed} Golbabee et al.,  have developed theoretical guarantees for the recovery of a matrix of rank $r$ and which has only $k$ non-zero rows using low rank and joint sparsity priors from its random Gaussian dense measurements. Unfortunately, the dense measurement scheme, where each measurement is a linear combination of all matrix entries is not practical in dynamic imaging; each measurement can only depend on a single column of the matrix. Another alternative is the multiple measurement vector scheme (MMV), where all the columns are measured by the same sampling operator \cite{chen2006theoretical}. This scheme offers a factor of two gain over the independent recovery of the columns, when the matrix is full rank; the gain is minimal when the rank of the matrix is far lower than the number of columns. This is clearly undesirable since the columns are highly redundant in the low-rank setting; one would expect significant gains in this case. 

We consider a two step strategy to recover a simultaneously low-rank and jointly sparse matrix from the measurements of its columns. Specifically, we propose to first recover the row subspace of the matrix from a set of common measurements made on the columns. Once the row subspace is estimated, the subspace aware recovery of the column subspace simplifies to a simple linear problem. This work is motivated by two-step algorithms used in dynamic MRI, where the temporal basis functions are first recovered from the central k-space samples \cite{liang}. While excellent reconstruction performance is reported in a range of dynamic and spectroscopic MRI applications  \cite{liang}, theoretical guarantees on the recovery of the matrix using this two-step strategy are lacking. A key difference of the proposed formulation with \cite{liang} is the assumption of joint sparsity, which plays a key role in ensuring perfect recovery. The joint sparsity of the matrix columns/ image frames is a reasonable assumption in dynamic imaging, where the image edge locations are approximately not changing from frame to frame . 

Our results show that the row subspace can be robustly recovered from a few measurements, which are common for all the columns. The number of  common measurements is dependent on the joint sparsity or rank, which ever is smaller. We also developed a sufficient condition to guarantee perfect subspace aware recovery of the matrix, once the row subspace is known. We verify the results using numerical simulations and demonstrate the utility of the scheme in recovering free breathing cardiac CINE MRI data. We observe that good recovery is possible when the number of measurements are comparable to the theoretical guarantees. We also observe that in addition to providing good guarantees on recovering the matrix, joint sparsity provides a significant improvement in performance in practical applications. 
 	
	\section{Proposed Approach} 
	
We consider the recovery of \textcolor{black}{ $\mathbf X \in \mathbb R^{n\times N}$} that is $k$-jointly sparse (has only $k$ non-zero rows) and has a rank of $r$ ($k$ and $r$ are independent). In the context of dynamic imaging, $n$ is the number of pixels in the image, while $N$ is the number of frames in the time series. The skinny singular value decomposition (SVD) of this matrix is specified by $\mathbf X =\mathbf U \boldsymbol\Sigma\mathbf V^H$, where the columns of $\mathbf U \in \mathbb R^{n\times r}$ and $\mathbf V \in \mathbb R^{N\times r}$ are orthonormal. We consider measurements that are only dependent on columns of the matrix, denoted by $\mathbf x_i$: 
	\begin{equation}
	\label{maineqns}
	\underbrace{\left[ \begin{array}{c} \mathbf z_{i} \\ \mathbf y_i   \end{array} \right ]}_{ \overline{\mathbf y}_i}=\underbrace{\left [ \begin{array}{c} \boldsymbol \Phi \\ \mathbf A_i    \end{array} \right ]}_{\mathbf D_{i}}~\mathbf x_{i}.
	\end{equation}
The measurement matrix $\mathbf \Phi \in  \mathbb C^{s \times n}$ is common for all columns, while different measurement matrices $\mathbf A_i$ are chosen for different columns. 

We introduce a two-step algorithm to recover the matrix from its measurements $ \overline{\mathbf y}_i;~i=0,..,N-1$. 
\begin{enumerate}
\item We show that the row subspace matrix $\mathbf Q=\mathbf R\mathbf V^H$ can be estimated from the common measurements $\mathbf Z = \boldsymbol \Phi \mathbf X$ as the eigen decomposition of $\mathbf Z^{H}\mathbf Z $. Here, $\mathbf R$ is an arbitrary invertible matrix, whose condition number is bounded under simple conditions on $\boldsymbol\Phi$. 
\item The subspace aware recovery of $\mathbf X = \mathbf P \mathbf Q^H$ in (1) simplifies to a linear system of equations. This system is invertible, if the matrix is $k$-jointly sparse and satisfies the condition ${\rm spark}(\mathbf X) = r+1$. The last sufficient condition implies that every $r$ columns of the matrix are linearly independent, which is a bit pessimistic. In reality, one requires considerably weaker conditions, which will be the focus of our future work.
\end{enumerate}
We will now derive conditions for the success of the above two steps.
	\subsection{Recovery of the row subspace}
The common measurements $\mathbf Z$ are related to the row subspace vectors $\mathbf V$ as 
\begin{equation}
\mathbf Z = \underbrace{\boldsymbol \Phi~\mathbf U~\boldsymbol \Sigma}_{\mathbf R}~ \mathbf V^H.
\end{equation}
We propose to estimate the subspace from the eigen decomposition of 
\begin{equation}
\mathbf Z^H \mathbf Z=\mathbf V\,\mathbf R^H\,\mathbf R\,\mathbf V^H.
\end{equation}
Note that if $\mathbf R$ is a full rank matrix, $\mathbf R^H\,\mathbf R$ is positive definite and has a singular value decomposition $\mathbf W \boldsymbol \Lambda \mathbf W^H$; where $\mathbf W \in \mathbb R^{r\times r}$ is an orthonormal matrix and all the diagonal entries of $\boldsymbol\Lambda$ are positive. Thus, the eigen decomposition of $\mathbf Z^H \mathbf Z$ yields 
\begin{equation}
\mathbf Z^H \mathbf Z=\left({\mathbf V \mathbf W}\right)\,\boldsymbol \Lambda\left({\mathbf V \mathbf W}\right)^H.
\end{equation}
Note that $\{{\rm span}(\mathbf w_i; i=0,..,r-1)\}=\{{\rm span}(\mathbf v_i; i=0,..,r-1)\}$ since $\mathbf {VW}$ is orthonormal. We now present conditions on $\boldsymbol\Phi$ that will guarantee $\mathbf R$ to be full rank.
	
	\begin{thm}
	The row subspace of $\mathbf X$ is uniquely recovered from the measurements $\mathbf Z = \boldsymbol \Phi \mathbf X$, if $\mathbf X$ is $k-$jointly sparse and iff ${\rm spark}(\mathbf \Phi)\geq k+1$.
	\end{thm}
	\noindent
We now show that the recovery of the subspace is also robust, when $\mathbf X$ is $k$-jointly sparse and the measurement matrix $\boldsymbol \Phi$ satisfies the restricted isometry property (RIP) for $k$ sparse vectors.
\begin{thm}
Suppose the measurement matrix $\mathbf \Phi$ satisfies the restricted isometry conditions for k-sparse vectors 
\begin{equation}
(1-\delta_k) \|\mathbf x\|_2^2 \leq \|\boldsymbol \Phi \mathbf x\|_2^2 \leq (1-\delta_k)\|\mathbf x\|_2^2
\end{equation}
then, the condition number of $\mathbf R, \boldsymbol \kappa$ is bounded by 
\begin{equation}
\kappa(\mathbf R) \leq \sqrt{\frac{1+\delta_k}{1-\delta_k}}~\kappa(\mathbf X)
\end{equation}
\end{thm}
The above conditions guarantee good recovery of the matrix when the measurement matrix $\mathbf \Phi$ satisfies the RIP conditions for $k$-sparse vectors. In many practical applications, the rank of $\mathbf X$ is much smaller than $k$. We now show that the row subspace can be reliably recovered using a $\mathbf \Phi$ with considerably lower number of measurements compared to $k$.

	\begin{thm}
	The row subspace $\mathbf Q$ of any matrix $\mathbf X$ can be uniquely recovered from the measurements $\mathbf Z = \boldsymbol \Phi \mathbf X$ for almost all matrices $\mathbf \Phi \in \mathbb C^{s \times n}$, if $s\geq r$.
\end{thm} 
The next theorem shows that $\mathbf \Phi \mathbf U$ is well conditioned, when $\boldsymbol \Phi$ has complex Gaussian random entries;  the condition number of $\mathbf R$ is bounded as long as $\mathbf X$ is well-conditioned.
\begin{thm}\cite[Theorem 3.2]{condition}
Suppose the entries of $\mathbf \Phi$ are independent, zero mean, complex Gaussian with unit variance.  Then for a constant M independent of c and for every $c>1$
\begin{equation}
Pr[ \boldsymbol \kappa(\mathbf \Phi \mathbf U)> c]  \leq  M c^{-2(s-r+1)}.
\end{equation}
\end{thm}
The constant $M$, defined in \cite{condition}, depends on $r$ and $s$ and is phrased as an expectation. Note that the probability that the condition number exceeds $c$ declines rapidly with a growing $c$, depending on $s-r+1$. The proofs will be added to a future work. 	
	
	
The above theorems guarantee the recovery of the row subspace of $\mathbf X$ from the common measurements of its columns, acquired by $\boldsymbol \Phi$. The number of common measurements depend upon the joint sparsity $k$ or the rank $r$, depending on which is smaller. In many dynamic imaging applications, $r<<k$ and hence the number of common measurements is dependent on the rank. This implies that very few common measurements are required to recover the subspace.

\begin{figure}[t!]
     \subfigure[{Proj. error vs common Gaussian samples}]{\includegraphics[width=.23\textwidth,height = 2.5 cm]{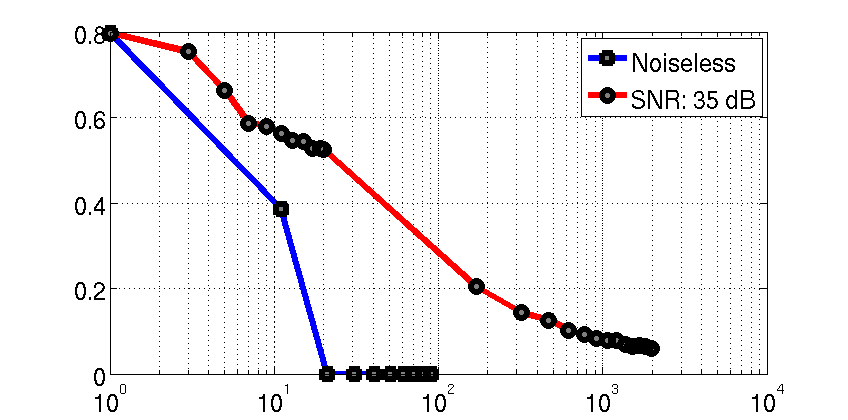}}
     \subfigure[{Proj. error vs common points on radial Fourier lines }]{\includegraphics[width=.23\textwidth,height = 2.5 cm]{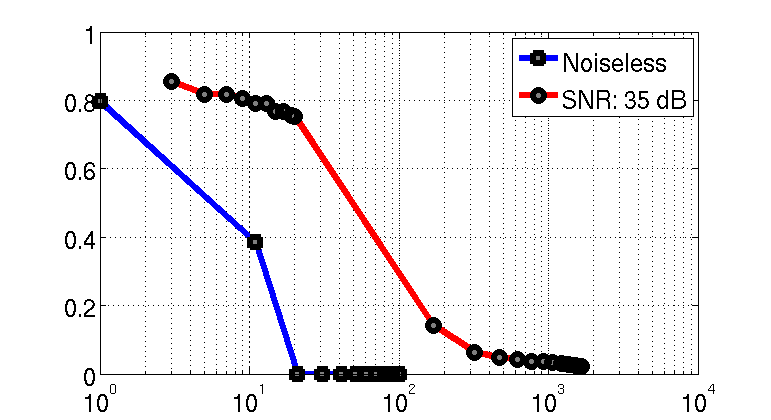}}
\caption{\small Projection error between subspaces vs \# common Gaussian samples (left) and common points on radial Fourier lines (right)} \vspace{-1.5em}
\label{x1rec}
\end{figure}

	\subsection{Subspace aware recovery of $\mathbf X$}
	Once \textcolor{black}{$\mathbf Q = \mathbf R \mathbf V^H \in \mathbb R^{N\times r}$} are obtained from the common measurements of the columns, the recovery of the matrix
\begin{equation}
\mathbf X = \underbrace{\mathbf U \boldsymbol \Sigma \mathbf R^{-1}}_{\mathbf P}\underbrace{\mathbf R \mathbf V^H}_{\mathbf Q^H}
\end{equation}
simplifies to the estimation of the coefficient matrix $\mathbf P \in  \mathbb C^{n \times r}$. Vectorizing both sides of second row of equation (2), we obtain
\begin{eqnarray}
\label{syseq}
	\underbrace{\left [ \begin{array}{c} \mathbf y_1  \\\vdots   \\ \mathbf y_N \end{array} \right]}_{{\rm vec}(\mathbf Y)}
	= \underbrace{\left [\begin{array}{ccc} q_{11}~\mathbf A_1 &\cdots& q_{r1}~\mathbf A_1 \\
	\vdots \\ q_{1N}~\mathbf A_N &\cdots& q_{rN}~\mathbf A_N \end{array} \right]}_{\mathbf B}
	\underbrace{\left [\begin{array}{c}\ \mathbf p_1 \\\vdots   \\ \mathbf p_r \end{array} \right]}_{{\rm vec}(\mathbf P)}
	\end{eqnarray}
Since $\mathbf X$ is jointly $k$ sparse, the sparsity of ${\rm vec}(\mathbf P)$ is $k r$.

We now introduce a sufficient condition 
\begin{equation}
\label{sparkcondition}
{\rm spark}(\mathbf X)=\mathbf r+1
\end{equation}
 to guarantee the recovery of $\mathbf P$ from (\ref{maineqns}). This condition implies that every collection of $r$ columns of $\mathbf X$ is linearly independent. In the absence of such a condition, there might exist columns of $\mathbf X$ that are orthogonal to all other columns of $\mathbf X$. To obtain perfect recovery of all the columns in this worst case scenario, we require ${\rm spark}(\mathbf A_i)=2k; \forall i=0,..,n$; there is no benefit over the independent recovery of the columns or the knowledge of the subspace. We now present a sufficient condition on the measurement matrices to guarantee the subspace aware recovery of $\mathbf X$ that is $k-$jointly sparse and has rank $r$, while satisfying (\ref{sparkcondition}).
\begin{thm}
Let $n=(p+1)r$, where $p$ is an arbitrary integer and the measurement matrices are chosen as
\begin{eqnarray}\nonumber
\mathbf C_1 & = &\mathbf A_1 = \mathbf A_2  .. = \mathbf A_r\\\nonumber
\vdots\\\label{clustering}
\mathbf C_{p} &=& \mathbf A_{pr+1} = \mathbf A_{pr+2}  .. = \mathbf A_{N} 
\end{eqnarray}
Here, $\mathbf C_i \in \mathbb R^{s_i \times n}; i = 1,.. p$. Then, $\mathbf P$ can be uniquely determined from (\ref{syseq}) if
\vspace{-2em}
\begin{equation}
{\rm spark}\left(\underbrace{\left[
\begin{array}{c}
\mathbf C_1\\
\vdots\\
\mathbf C_p\\
\end{array}
\right]}_{\mathbf C}
\right) \geq 2k.
\end{equation}
\end{thm}

  

\begin{SCfigure}
  \includegraphics[width=0.24\textwidth]{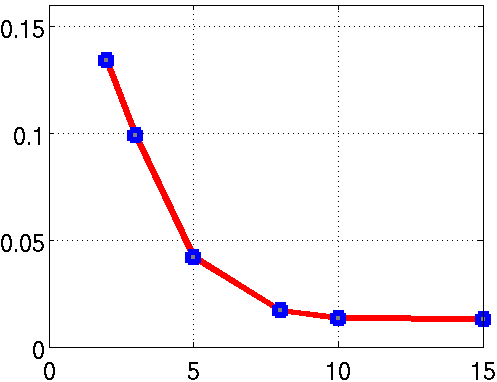}
  \caption{Recovery error vs \# variable radial lines} 
    \label{x1rec}
\end{SCfigure}

The classical MMV scheme requires a total of $(2k-r+1)N$ measurements for its unique recovery of a matrix of dimension $n \times N$ and rank $r$. The total number of measurements required by the dense measurement scheme is considerably lower and of the order of the degrees of freedom in a matrix \cite{golbabaee2012compressed}. Combining the results in the above subsections, the proposed scheme requires of the order of $(2k-r+N)r$ for unique recovery---or equivalently $r + 2kr/N$ measurements per frame; this is comparable to the degrees of freedom in the matrix and is comparable to the best possible scenario involving dense measurement matrices. Considering that the dense measurement scheme is impractical in a dynamic imaging setting, the gains offered by the practical efficient two step strategy is quite significant.
	
\subsection{Algorithm}
We pose the recovery of the jointly sparse vector $\mathbf P$ from the linear measurements (\ref{syseq}) as a $\ell_1$ minimization scheme:
\begin{equation}
\hat{\mathbf P} = \arg \min_{\mathbf P} ~ ||\mathbf B\, {\rm vec}(\mathbf P) -{\rm vec} (\mathbf Y)||_2 ^2 + \boldsymbol \|\mathbf T \mathbf P\|_{\ell_1-\ell_2}
\end{equation}	 
Here, $\mathbf T$ is an appropriately chosen transform or frame operator, while $\ell_1-\ell_2$ norm is the mixed norm to encourage joint sparsity. In this work, we use $\mathbf T$ as the finite difference operator. We solve the above problem using the alternating direction method of multipliers (ADMM) algorithm.\cite{yang2009alternating}. 

\section{Results}
We first validate our results using numerical simulations on PINCAT phantom corresponding to CINE MRI data, before using the framework to recover free breathing CINE data. 
\subsection{Numerical simulations}
We consider a PINCAT phantom with dimension of 128 x 128 x 200 and a rank of 20. In this case, the rank $r$ is far less than sparsity $k$. We first determine the accuracy of the subspace matrix, recovered from the common lines. We use the projection error between two subspaces $\mathbf V_1$ and $\mathbf V_2$ is defined as
\begin{equation}
\mathcal E = \frac{||(\mathbf I-\mathbf V_1\mathbf V_1^H)\mathbf V_2||_2^2 +||(\mathbf I-\mathbf V_2\mathbf V_2^H)\mathbf V_1||_2^2 }{||\mathbf V_1||_2^2+||\mathbf V_2||_2^2}.
\end{equation}
as the metric for comparing two subspaces. In Fig. 1 we plot the projection error vs the number of common Gaussian samples (left) and common points on radial Fourier measurements(right). Noiseless and a noisy setting with a SNR of 35 dB are compared. We observe that the projection error drops to zero when the number of samples equals the rank 20 in the noiseless cases. We also observe that good estimates for the subspaces can be obtained with more measurements in the noisy setting, indicating that the recovery is robust to noise. 

\begin{figure}
\begin{center}\includegraphics[width=.4\textwidth]{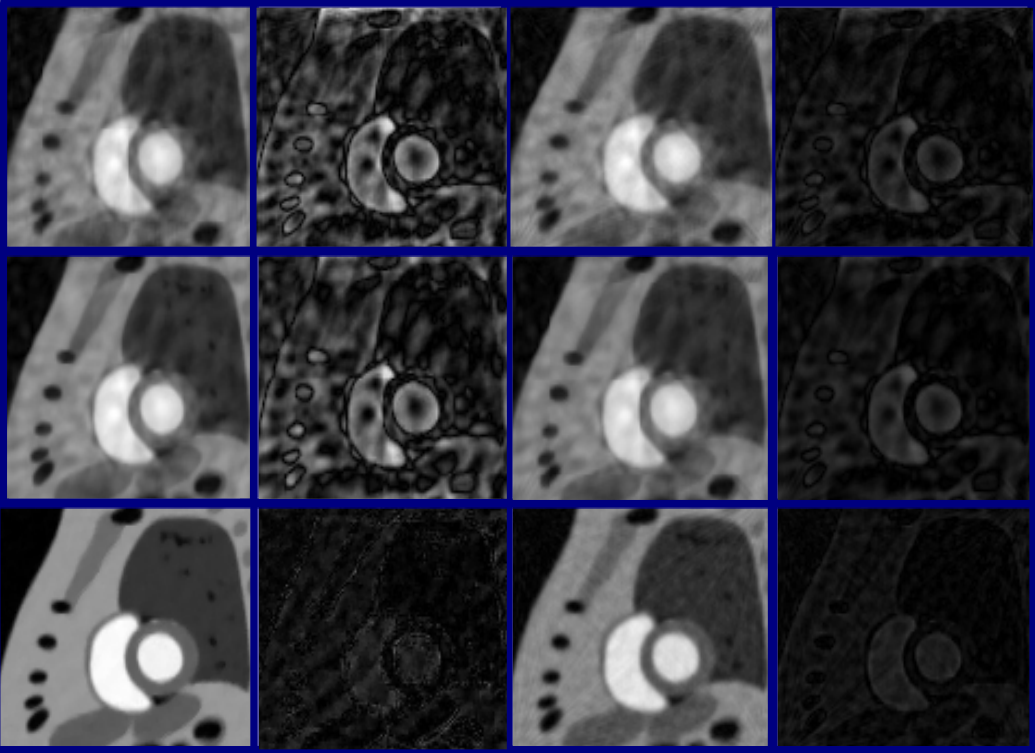}\end{center}
\caption{\small Reconstructed Pincat phantom. Top: No regularization, Middle: Standard TV regularization, Bottom: Joint sparsity regularized, Noiseless reconstruction and error images on first two columns and the corresponding noisy (SNR of 50 dB) on the last two columns } \vspace{-1.0em}
\label{x1rec}
\end{figure}

		  In Fig. 2, we consider the subspace aware recovery of the matrix using the subspace estimated from \textcolor{black}{4  common radial lines}. We recovered the images using joint sparse TV recovery. The normalized recovery error as a function of the number of radial lines used in each frame. We observe that we obtain a recovery error of 1\% when eight radial lines/frame are used; this corresponds to an acceleration of approximately 10.7. We  expected the error goes down with more number of lines. We show the reconstructions corresponding to \textcolor{black}{4 common radial lines and 5 variable lines} in Fig. 3. The rows in Fig. 3 corresponds to the reconstructions obtained when $\mathbf P$ is recovered with no regularization, standard spatial TV regularization,  and the proposed joint sparsity regularization. The first two columns show the reconstructed image and the error image w.r.t the original phantom in the noiseless case. The corresponding noisy cases are shown in the last two columns with an output SNR of 50 dB.

\subsection{Recovery of free breathing cardiac CINE data}	  
We demonstrate the utility of the algorithm in recovering free breathing CINE data in Fig. 4. The data was acquired using an SSFP sequence with an $18$ channel coil array, with TR/TE of $4.2/2.1$ ms, matrix size of $512\times512$, FOV of $300$mm$\times300$mm and slice thickness of $5$mm on 3T Siemens Trio scanner.  We considered $12$ radial lines of k-space to reconstruct each image frame, $4$ of which were common lines. This translated to a temporal resolution of $50$ ms. The acquisition time was $25$ s which corresponds to $500$ image frames. The rows correspond the the reconstructions obtained when $\mathbf P$ is recovered with no regularization, standard TV regularization and the proposed joint sparsity regularization. The last column shows the time profile along a vertical line. The results show the utility of the proposed scheme in providing good reconstruction of free breathing CINE MRI data. 
 
	\section{Conclusion}
	We introduced a two step algorithm with recovery guarantees to reconstruct a low rank and jointly sparse matrix from its under sampled measurements. The results show that under simple assumptions, the two step recovery scheme is guaranteed to provide good recovery of the matrix. The application of the scheme to the recovery free breathing CINE data demonstrates the utility of the scheme in practical applications.
	
\begin{figure}
\includegraphics[width=.5\textwidth]{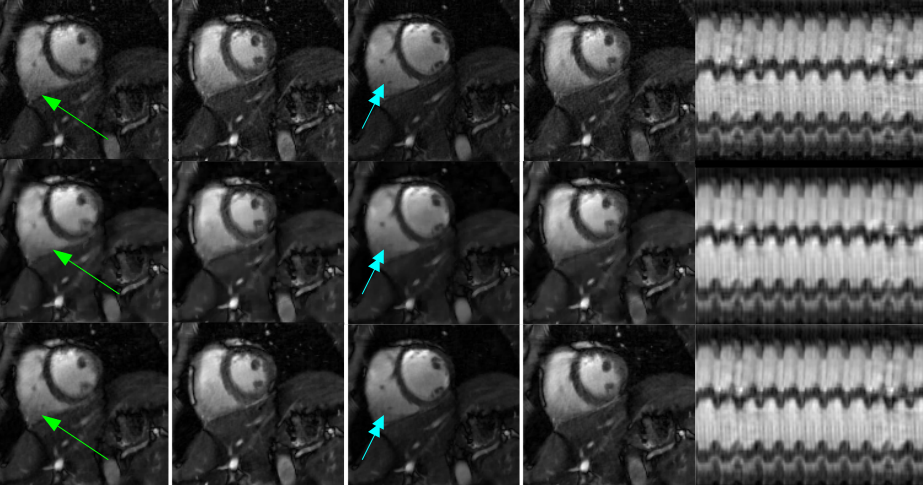}
\caption{\small Reconstructed free breathing CINE data. Top: No regularization, Middle: Standard TV regularization, Bottom: Joint sparsity regularized. Last column shows the time profile along the myocardium.} \vspace{-1.0em}
\label{x1rec}
\end{figure}

\bibliographystyle{IEEEtran}
\bibliography{refs.bib}
	 
	\end{document}